\documentclass[11pt,letterpaper]{article}
\usepackage{naaclhlt2016}
\usepackage{times}
\usepackage{latexsym}
\usepackage[utf8]{inputenc}
\usepackage{amsmath}
\usepackage{amsfonts}
\usepackage{amssymb}
\usepackage{bm}
\usepackage{xcolor}
\usepackage{graphicx}
\usepackage{subfig}
\usepackage{float}
\usepackage{enumitem}
\usepackage{url}
\usepackage[title]{appendix}
\usepackage{afterpage}

\naaclfinalcopy 

\title{Detecting cities in aerial night-time images by learning structural invariants using single reference augmentation\bigskip\\ \large Project in\\\textit{Image Classification}\\Winter 2018\\Prof. Dr. Tatjana Scheffler\\}

\author{Philipp Sadler\\University of Potsdam}

\begin{document}

\maketitle

\begin{abstract}
	This paper examines, if it is possible to learn structural invariants of city images by using only a single reference picture when producing transformations along the variants in the dataset. Previous work explored the problem of learning from only a few examples and showed that data augmentation techniques benefit performance and generalization for machine learning approaches. First a principal component analysis in conjunction with a Fourier transform is trained on a single reference augmentation training dataset using the city images. Secondly a convolutional neural network is trained on a similar dataset with more samples. The findings are that the convolutional neural network is capable of finding images of the same category whereas the applied principal component analysis in conjunction with a Fourier transform failed to solve this task. 
\end{abstract}

\section{Introduction} 

Sanchez de Miguel analysed light pollution on a global scale and proclaimed along his publication the \textit{Cities at Night} project \cite{sanchez}. After starting the initiative, a global night-time map showing the light pollution at night on the whole planet has been compiled. 

For this purpose hundreds of images taken from the \textit{International Space Station (ISS)} were evaluated. Nevertheless, the cities evolve over time, so that the light pollution of each specific city is changing. Such adjustments would need an ongoing evaluation of aerial night-time images which is a cumbersome and cost intensive work. 

As these cities are expected to keep their overall structure, is there a method to detect specific cities within such images? Furthermore, as the already established dataset is relatively small for modern machine learning approaches with only a handful of samples per city: 

Is it possible to learn structural invariants of the cites by using only a single image of a city when producing transformations along the variants in the dataset?

\section{Related Work}

\cite{Fei-fei06one-shotlearning} explored the problem of learning from only a few examples using a Bayesian approach and worked out that object detection with just a few or handful of images is possible by transferring knowledge from previously learned categories. The idea of single example learning has also been explored in a synthetic setting for character classification in combination with nearest neighbours \cite{NIPS2004_2576}. One of the latest approaches incorporates a deep learning technique called Neural Turing Machines \cite{DBLP:SantoroBBWL16}. In this paper a machine learning approach with a Fourier Transform feature extractor in conjunction with a principal component analysis is examined as a baseline for image classification using only a single city image. Then a deep learning approach with convolutional neural networks is applied for this problem. 

Since most of the city variants within the \textit{Cities at Night} dataset are known, the single sample learning is combined with data warping techniques. \cite{DBLP:WongGSM16} came to the conclusion that data warping shows benefits in performance and generalization for machine learning approaches on a hand-written digits dataset (MNIST). One of the latest approaches compares this traditional data transformations like shrinking, stretching, shifting and rotating with a method called neural augmentation that learns the best data transformations for a neural network using Generative Adversarial Nets \cite{DBLP:Perez}. Nevertheless, in this paper only traditional data transformations are applied to the single reference image, which should produce variants along the underlying manifolds of the aerial night-time images.

\section{Cities at Night Dataset}

The cities at night team has assembled a list of 3625 images taken from the International Space Station during night phase of the overflight \cite{pmission}. For each image \textit{id}, \textit{mission} and \textit{coordinates} are listed. Given these attributes 3126 images in small resolution could be downloaded automatically from the NASA project hosting the images \cite{nasa}. The rest of the images was not retrievable anymore.

\subsection{Quality and Variations}

\begin{figure}[h]
	\centering
	\subfloat
	[The moving camera results into smaller and larger shifts within the image during a shot series. Here an extreme example is shown.]
	{
		\includegraphics[width=0.45\textwidth]{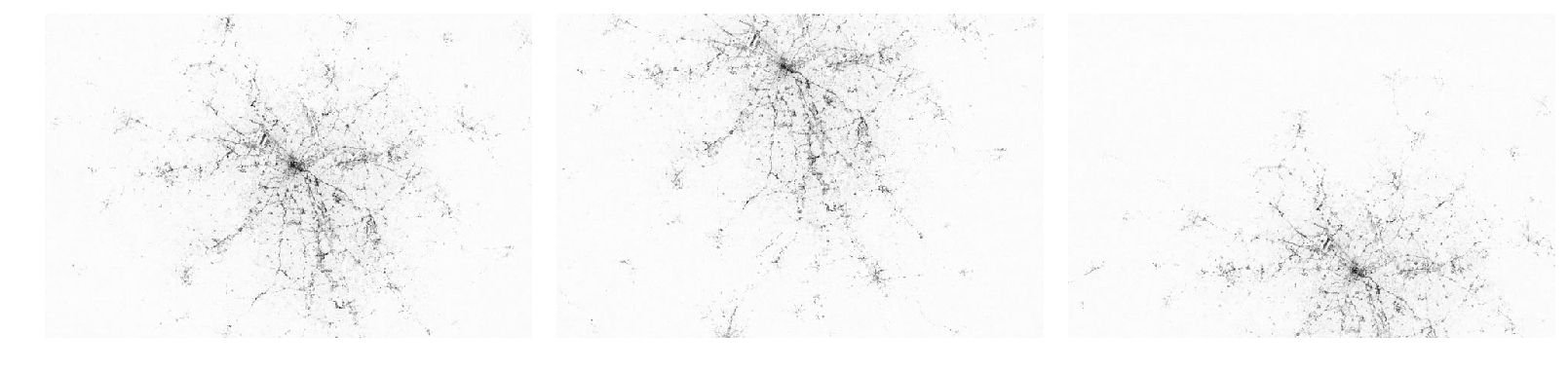}
		\label{fig:shift}
	}
	\hfill
	\subfloat
	[In addition, the movement causes small rotations to the images and sometimes the series are applied with different zoom levels.]
	{
		\includegraphics[width=0.45\textwidth]{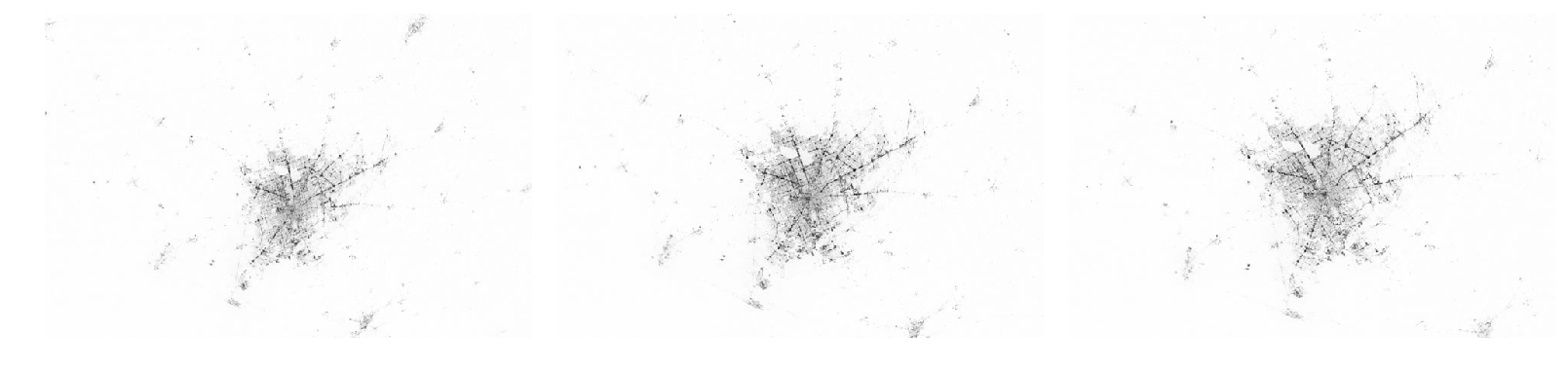}
		\label{fig:rotation}
	}
	\hfill
	\subfloat
	[There are also always small differences in pixel intensity for the same spot or the series had been taken intentionally with different periods of exposure.]
	{
		\includegraphics[width=0.45\textwidth]{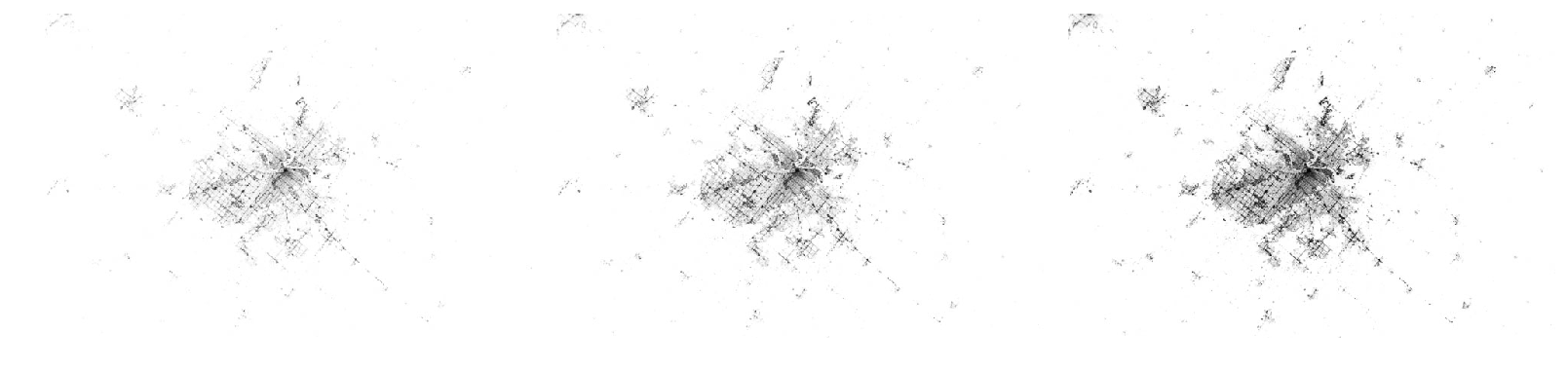}
		\label{fig:brightness}
	}
	\caption{The quality of the images within the Cities at Night dataset is highly varying as shown here example-wise. The sample images are in reverse grey-scale for better visibility of the distortions.}
	\label{fig:variations}
\end{figure}

The small resolution images are of size 640x426 pixels (width x heigth), between 240 and 300 dpi and in focal lengths between 50mm and 400mm. As these images are taken mainly occasional and yet for documentation purposes only, the dataset is quite noisy in respect to the task of elaborating them as cities at night.  There are images showing the earth ellipse, multiple cities, parts of station equipment or that are blurry because of wet lenses (Appendix \ref{appendix:quality}). In addition some images are damaged or black only. Aside of this noise, there are also a lot of good single city images within the dataset. Nevertheless, these good shots show variations themselves because of the moving camera and the series-like nature of the shots as shown in Figure \ref{fig:variations}

\subsection{City Subsets}

Subsets of images for specific cities can be extracted from the dataset by linking the pmission map view with its table representation. The image list is then restricted to the data points shown in the map view, so that a subset for Berlin, Paris and Madrid is retrievable. From these retrieved subsets two Berlin, three Paris, two Madrid images are removed, because they are probably mislabelled, resulting into 12 Berlin, 14 Paris and 31 Madrid images for each subset. These cities are taken as samples because for a human classifier Berlin and Paris look rather similar, but Madrid looks different from both. The learning algorithm should show capabilities to distinguish similar and different looking cities.

\section{Single Reference Augmentation}
\label{section:dataset}

Modern iterative image classifications algorithms based on neural networks need usually thousands of good samples for a single category to learn good representations for this class. In contrast to that existing datasets are often quite small because the data is often time-consuming, expensive or risky to collect. The \textit{Cities at Night} dataset is time-consuming and expensive to collect, because the images are taken from the ISS, so there is only a very limited number of samples for each category (with below 100 images per city). 

Nevertheless, there exist already labelled images for the fourteen cities Berlin, Calgary, Frankfurt, Lisbon, Nantes, Paris, Perth, Saskatoon, Seattle, Tacoma, Tehran, Toronto, Vancouver and Winnipeg. Furthermore, the city images are expected to be structurally the same over time, but are only varying because of the circumstances when taking the pictures. Given these preconditions, the single reference augmentation technique should take only a single labelled city image and reproduce images that show the variations caused by the photo shooting procedure. 

\begin{figure}[h]
	\centering
	\subfloat[The original single reference images for Berlin, Madrid and Paris.]
	{
		\includegraphics[width=0.5\textwidth]{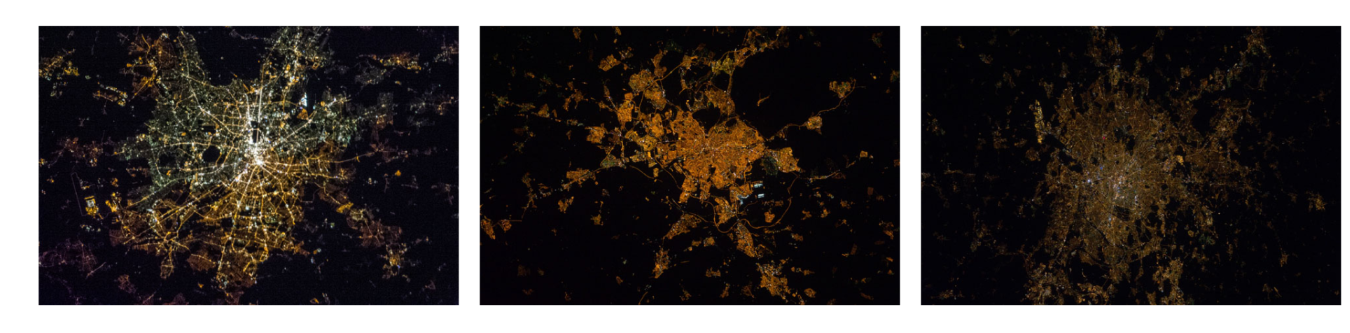}
	}
	\hfill
	\subfloat[The grey-scaled images of Berlin, Madrid and Paris.]
	{
		\includegraphics[width=0.5\textwidth]{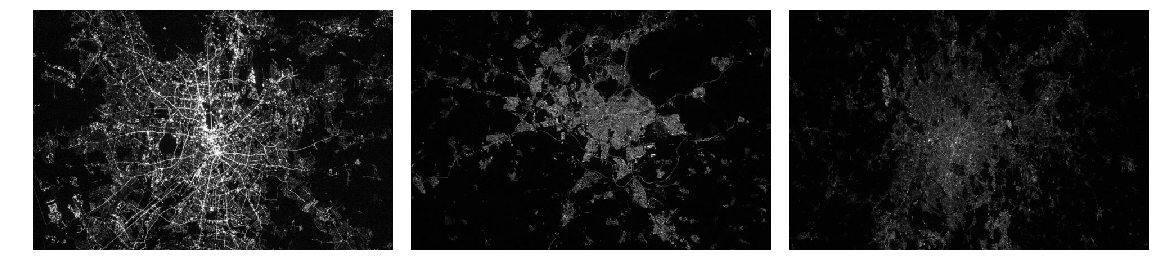}
	}
	\hfill
	\subfloat[The same images after applying the intensity rescale.]
	{
		\includegraphics[width=0.5\textwidth]{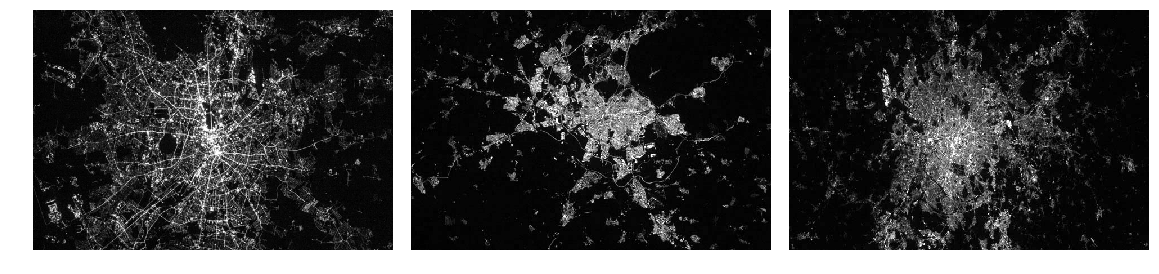}
	}
	\hfill
	\subfloat[The same images after enhancing and resizing.]
	{
		\includegraphics[width=0.5\textwidth]{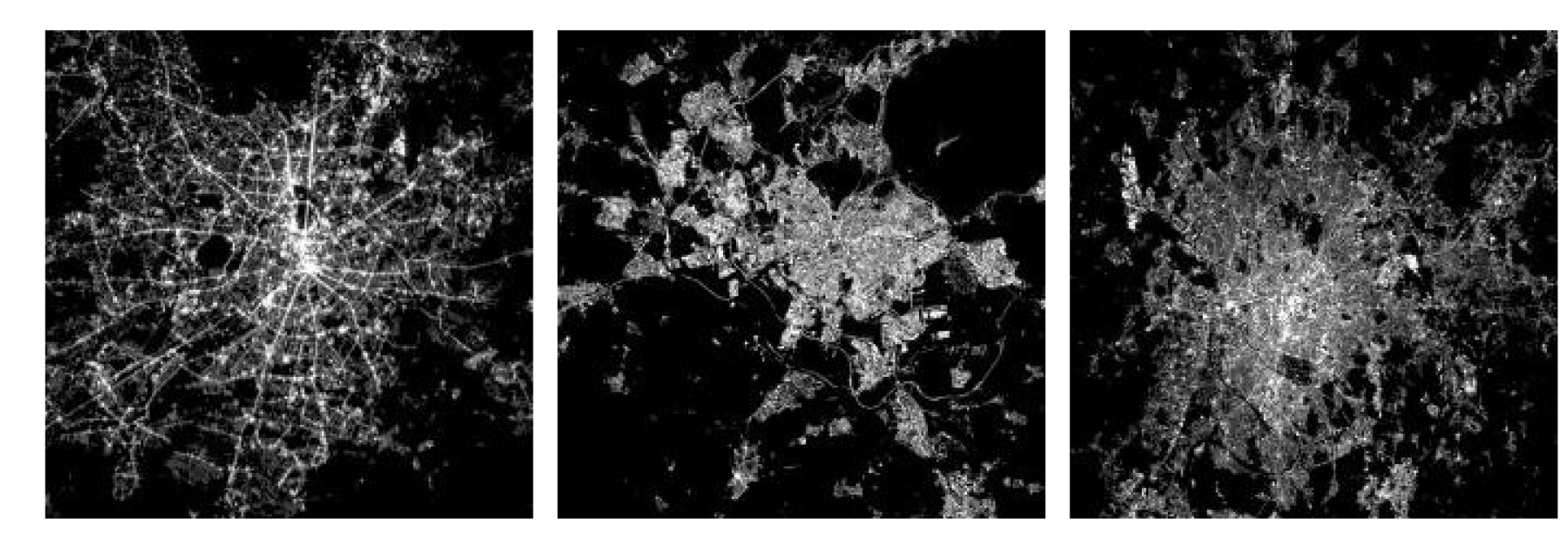}
	}
	\caption{The image enhancement and resizing steps for the single references of Berlin, Madrid and Paris from (a) to (d). The original images are prepared for the learning algorithm while for a human observer significant details are kept.}
	\label{fig:preprocessing}
\end{figure}

\begin{figure*}[h]
	\centering
	\subfloat[Transformations of the enhanced Berlin reference image.]
	{
		\includegraphics[width=0.6\textwidth]{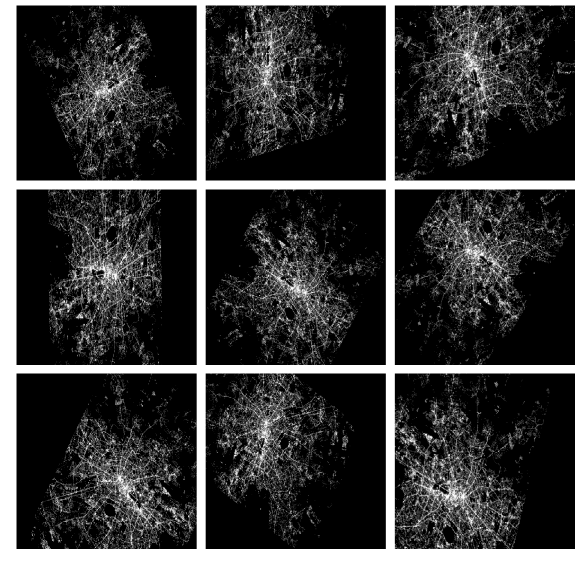}
	}
	\hfill
	\subfloat[The resized and cropped transformations of the same images.]
	{
		\includegraphics[width=0.6\textwidth]{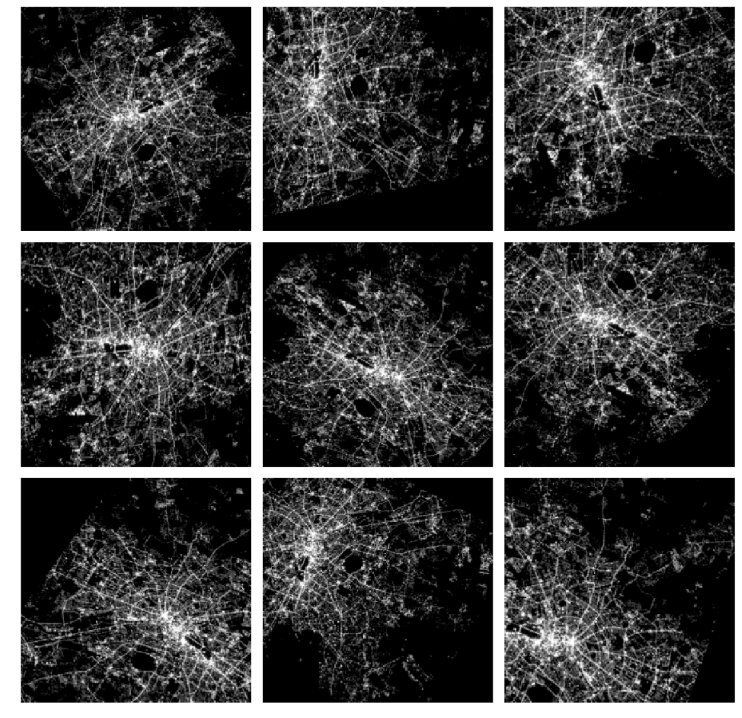}
	}
	\caption{(a) A single Berlin sample image has been enhanced and referenced for transformation. (b) Then the resulting variants of the reference image have been resized and cropped.}
	\label{fig:augmentations}
\end{figure*}

The algorithm is then expected to learn the image variations or find invariants for the specific cities within the pictures. When using single reference augmentation to create a dataset by performing the following steps as shown in Figure \ref{fig:preprocessing}, then exemplary results as in Figure \ref{fig:augmentations} are produced:

\begin{enumerate}
	\item \textbf{Perform enhancements} with grey-scale, better contrast and a pixel value threshold.
	\item \textbf{Perform transformations} with rotating, shifting, shearing, zooming and flipping.
	\item \textbf{Perform resizing} with crop to 426x426 pixels, rescaling to 256x256 pixels and central cropping to 224x224 pixels.
\end{enumerate}

\subsection{Convert to Grey-scale}

The images are converted from 3-dimensional RGB colours to 1-dimensional grey-scale. The reason for this is that the structural part of the cities within night-time pictures is highly encoded in the brightness of pixels, but not in the colours. Although this removes information from the images, this is expected to prevent learning algorithms from relying on colour values, but focus on the structural parts of the image. For example an algorithm that has been only shown red cars would easily predict a car for an image, only based on the fact that the picture is containing red coloured pixels and not because of the shape of the  content. In such a way, grey-scaling should reduce side effects when learning from the images. Apart from that, it is computational more expensive to rely on 3 channels when they are not necessarily required.

\subsection{Rescale Pixel Intensity}

After grey-scaling the images, pixel intensity values are rescaled between the percentiles 0.2 and 0.998. The structural properties of the image are made more explicit by enhancing the contrast in this way. The expectation is that lighter/darker pixels become more equally important for the learning algorithm.

\subsection{Threshold Pixel Values}

\begin{table}
	\centering
	\small
	\text{Taking the \textbf{mean} as threshold: }
	\begin{tabular}{|  l | r | r | r | r | }
		\hline
		& Th. & Bef. & Incr. & Spars.\\
		\hline
		Paris    & 38&    1291 & 182817 & 0.67\\
		Berlin   & 26&    5038 & 211977 & 0.78\\
		Madrid   & 29&     8886 & 186096 & 0.68\\
		\hline
	\end{tabular}
	\vskip .5em
	\text{Taking the \textbf{median} as threshold: }
	\begin{tabular}{| l | r | r | r | r | }
		\hline
		& Th. & Bef. & Incr. & Spars.\\
		\hline
		Paris     & 17 &  1291 & 132169 & 0.48\\
		Berlin    &  5 &   5038 &  66403 & 0.24\\
		Madrid     &  7 &  8886 & 103031 & 0.38\\
		\hline
	\end{tabular}
	\vskip .5em
	\text{Taking the \textbf{.25 percentile} as threshold: }
	\begin{tabular}{| l | r | r | r | r | }
		\hline
		& Th. & Bef. & Incr. & Spars.\\
		\hline
		Berlin  & 7  &  1291 &  52228 & 0.19\\
		Madrid  & 3  &  5038 &  13318 & 0.05\\
		Paris   & 4  &  8886 &  25416 & 0.09\\	
		\hline
	\end{tabular}
	\caption{Threshold methods on a single sample of Berlin, Madrid and Paris. For example, the mean threshold value 26 for Madrid leads to 211,977 additional zeros resulting into total 217,015 zero pixels for that image. This is a sparsity level of 78\% based on a resolution of 426x640 (272,640 pixels).}
	\label{table:thresholds}
\end{table}

Image pixel values below a certain threshold value are set to zero, so that pixel noise is reduced. For example dark parts within a city picture look black for a human observer, but still have small light intensities. The application of a pixel values threshold is expected to prevent a learning algorithm to make use of these unnoticeable pixel values. Furthermore, the transformations like shifts and rotations are producing additional space at the sides and corner of an image. This space is filled with zero valued pixels. As zero light intensities naturally not exist in pictures taken with a camera, the learning algorithm could easily learn that images containing zero-pixels are transformations and not originals. Taking a pixel threshold should eliminate this effect.

\newpage

For the three threshold methods \textit{mean}, \textit{median} and \textit{.25 percentile}, the most sparsity is achieved with the \textit{mean} threshold as shown in Table \ref{table:thresholds} while containing visual stability of the image. 

\subsection{Transformations}

The central idea of single reference augmentation is to produce variants of images that show variations from the set of variations that is existing in the dataset. The variants are produced with a combination of the following transformations:
\begin{itemize}
	\item rotation up to 180 degree
	\item width shift up to 20 percent
	\item height shift up to 20 percent
	\item shear range up to 20 percent
	\item zoom range up to 20 percent
	\item vertical and horizontal flip
\end{itemize}

The transformations produce unoccupied space within the images which is filled with zero valued pixels.

\subsection{Rescaling and Cropping}

The original images come with a resolution of 640x426 pixels which result into 272,640 pixels of which each is representing a single feature of the image. As classification outcomes should not rely on specific per-pixel details, but greater structures, it is common to reduce the dimensionality of feature space by rescaling and cropping.

\cite{DBLP:ZeilerF14} describes a method in which the smallest image dimension is resized to 256 pixels and then the image is cropped to 256x256 pixels. After that, usually tiles of shape 224x224 are taken from the crop e.g. 4 corner tiles and a center tile. As a result, a dataset is easily enlarged by the factor of 5. 

In deviation from that, the intention in this paper is to perform augmentation anyway, so the additional tiles are not necessary, because the augmentation produces already additional samples. Thus here only the rescaling and cropping is performed accordingly:

\newpage

\begin{enumerate}
	\item The images of 640x426 pixel are cropped to 426x426 pixel by taking away 107 pixel on both sides of the width. The image is not resized directly to 256x256 pixel, so that more details are kept.
	\item The crops of 426x426 pixel are then rescaled to a resolution of 256x256 pixel with application of a bilinear filter. Then again a central crop of 224x224 pixel is taken to keep a better resolution from the sample than just resizing to the target size. 
\end{enumerate}

\clearpage

\clearpage

\section{Experiments}

A useful application would be to detect images in the \textit{Cities at Night} dataset that show the overall structure of single reference city images with high precision and confidence so that manual work for filtering these images is limited. These found cities are then automatically processable in a tooling environment.

The use of only a single reference image within a training dataset might easily lead the classification algorithm to overfit the training data. Therefore a nice, but harder to accomplish property would be to find not just variants of the reference image, but also to generalize beyond the transformations and detect images that show details or context of the reference image. These found new variants could then be used for further class training. 

The experiments should show how well an image classification algorithm can be trained on a \textit{Single Reference Augmentation} dataset to variants of the same class within a noisy test dataset. For this purpose a single sample for the classes Berlin, Madrid and Paris is selected and based on these samples an according training dataset is created.

In the first experiment, a principal component analysis in conjunction with a Fourier transform is trained on a single reference augmentation training dataset using the city images. Then in a second experiment, a convolutional neural network is trained on a similar dataset with more samples. Both approaches are expected to find images of the same city within a noisy and unclean test dataset.

\subsection{Principal Component Analysis with Fourier Transforms}

The Principal Component Analysis (PCA) is a common baseline approach for image classification tasks. The PCA algorithm tries to find components within the data that describe most of its variance. This algorithm has been successfully applied to a Face recognition task using Eigenfaces \cite{DBLP:TurkP91}. The resulting components for face pictures are called Eigenfaces. When this Eigenfaces are learned, then every face image can be constructed in terms of these representations. The feature space of the face image has been transformed and reduced from pixel values to its Eigenfaces. 

In analogy to that method, the PCA should find Eigencities of the given city images. In addition, this experiment uses a 2D Fourier transform (FT) as a feature extractor right before the component analysis. The FT has already been successfully used to enhance face recognition tasks with PCA \cite{Bouzalmat}. Furthermore, the expectation is that the FT provides some stability across the image variants.

\subsubsection{Training}

The training dataset for this experiment consists of 100 single reference augmentation samples for each of the three classes \textit{Berlin, Madrid and Paris}. The resulting 300 images of shape 224x224 are preprocessed as described in Section \ref{section:dataset}.

The first operation applied on the dataset is a 2D-FT. Then each of the resulting 2D images are unrolled into a single column vector. Furthermore, the matrix of samples is standard-scaled\footnote{Substracted mean and divided by standard deviation} because the PCA is working best, when the data shows standard normalized distribution. Finally, six principal components are computed for the matrix of samples using PCA. More than six components do not show reasonable significance.

\subsubsection{Evaluation}

For the test dataset 10 images of \textit{Berlin, Madrid, Paris and Other} have been selected. The test dataset does not contain transformations, but preprocessed originals of the \textit{Cities at Night} dataset.

Analogous to the training, a 2D-FT is applied on the images as a first step. Then the images are unrolled and standard-scaled with the metrics of the training dataset. Finally, the samples are transformed to the Eigencity values using the learned principal components during training. The resulting dimensionality reduced representations of the images in terms of Eigencities are compared pairwise with cosine distance.\footnote{Cosine distance values are between $[0,2]$ with 0: totally similar, 1: orthogonal, and 2: opposite direction} Therefore each of the 40 test set result vectors is compared with each of the 300 training set result vectors.

A validation image is labelled with the training class that satisfies the most entries under a certain threshold. For example, if 50/100 Berlin, 20/100 Paris and 30/100 Madrid training samples have a cosine distance below a threshold of 0.5 with a validation image, then the Berlin class is chosen for the validation image. If there are multiple classes with the same amount of entries then the one with the best similarity score is chosen. The evaluation has been performed with all thresholds in $[0,1]$ using a step-size of 0.05.

\subsubsection{Results}

\begin{figure*}[t]
		\centering
		\includegraphics[draft=false,width=\textwidth]{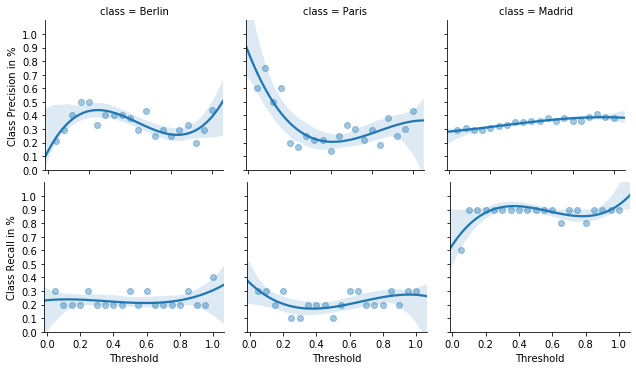}
		\caption{Precision and recall for thresholds between $[0,1]$ per class Berlin, Madrid or Paris. The prediction has been applied on the test dataset with 40 images of which 10 belong to each class additionally Other. A 3-order polynom is fit through the data points.}	
		\label{fig:pca_per_class}
	\end{figure*}

The evaluation results show for all thresholds a precision less equal than 50\%. As drawn in Figure \ref{fig:pca_per_class}, the precision for Berlin is highest with 50\% around a threshold of 0.25 while for Paris the precision is highest around 60\% for thresholds below 0.25. The precision for Madrid is almost the same with around 30\% along all thresholds. Accordingly Madrid shows constantly high values around 90\% for class recall over all thresholds. Whereas the for Berlin and Madrid only values around 30\% are achieved.

\begin{figure*}[h]	
		\centering
		\subfloat
		{
			\includegraphics[draft=false,width=0.45\textwidth]{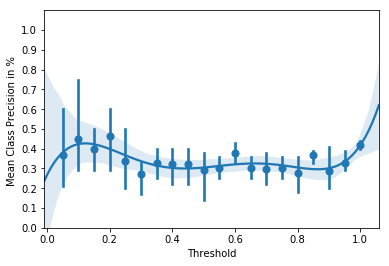}
		}
		\hfill
		\subfloat
		{
			\includegraphics[draft=false,width=0.45\textwidth]{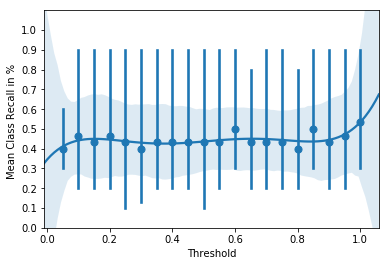}
		}
		\caption{The mean precision for thresholds between $[0,1]$ for the three classes Berlin, Madrid and Paris based on the data points in Figure \ref{fig:pca_per_class}.}	
		\label{fig:pca_mean}
	\end{figure*}

The results plotted in Figure \ref{fig:pca_mean} underline the per-class observation as the mean precision for all classes is always less equal than 50\% and the highest around a very restrictive threshold of 0.1. The same is shown for mean recall, but with a higher variance in the data points, because Madrid is always showing high values and the other classes low ones.

\subsubsection{Discussion}

The high values of recall while containing low precision for Madrid in contrast to Berlin and Paris show that the trained algorithm favours Madrid predictions beyond Berlin and Paris ones. Thus the algorithm shows same behaviour as when trained with an additional Other class. 

Previous experiments have shown that training with an Other class would lead the algorithm to predict most of the times Other. Therefore Other was excluded from the training set. The Madrid and Other training images seem to capture properties that are distributed along all images, so that the similarity scores are better for these images. 

The reasons for this cannot be pinpointed here, but usually PCA is not robust to unaligned samples and background changes. Therefore maybe the training data for Madrid is better aligned than the other classes or the Fourier transform as a feature extractor is not working as expected. Solutions to this problem might be to find a method to centralize the cities on the images after transforming them and to find a better method to clean the background instead of just filling empty space with zeros.

In any case, the single reference augmentation method is based on learning these transformations. Therefore the natural choice would be to enhance the training dataset so that more training samples per class could be compared to the validation images. This would lead to huge matrices which can be overcome only by an incremental learning algorithm which complicates the PCA training.

To conclude, the detection of cities in night-time images with Fourier transform and Principal Component Analysis is not comparable with the successful application for Face detection using Eigenfaces, because the cities show more transformations and variants. It would be more comparable with detecting the same face transformed in many ways. As it seems there are no useful reductions to Eigencities which satisfy the variants, but they are rather similar which results in bad classification results.

\afterpage{\clearpage}
\afterpage{\clearpage}

\subsection{Convolution Neural Network}

Nowadays, Convolutional Neural Networks (CNN) have become the de-facto-standard for object detection and image classification. Here multiple kernels are convolved with the input to learn specific patterns of the underlying distribution. A widely known successful application of this technique was already published in \cite{Lecun} for recognition of handwritten numbers and text. 

In analogy to that application, the CNN in this paper should learn lower and higher level patterns of the specific cities using only a single reference image for each city. This could be successful, because in contrast to the hand-written number problem which show numbers depending on the handwriting, the city structures themselves are largely invariant, but transformed with shifts, rotations, zooms and shears. 

The CNN is expected to learn the invariant structures e.g. overall city street formations or specific lighter and darker spots such as airports and parks, while the single reference augmentation should help to prevent overfitting on specific images of the city and encourage the network to become more invariant to the transformations.

\begin{figure*}[p]
	\centering
	\includegraphics[width=1.0\textwidth]{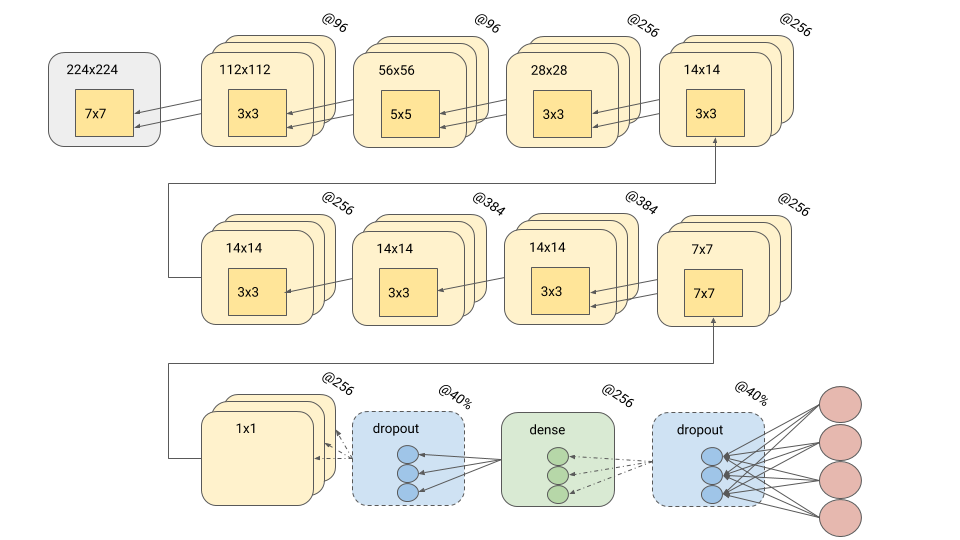}
	\caption{An architectural view of the applied convolutional neural network. The rectangular boxes are the layers containing the layer size: The grey one represents the input images, the yellow ones depict convolutional layers, the blue ones depict optional dropouts layer, the green ones are dense layers and the red circles are the output classes. The amount of arrows is indicating the stride e.g. 1 arrow is stride 1. The small boxes represent the kernels and their sizes. The \textit{@N} indicate the amount of feature maps for convolutional layers, the amount of units for dense layers and the change of dropout for dropout layers.}
	\label{fig:architecture}
\end{figure*}

The CNN applied in this paper is inspired by \cite{DBLP:ZeilerF14}. As depicted in Figure \ref{fig:architecture} the network consists of $12+2$ layers. In addition to the original 12 layers, two optional dropout layers with 40\% are added after each of the two fully connected ones. Dropout is a useful and easy way to push the network towards a more generalized solution \cite{DBLP:KuboTW16} \cite{JMLR:srivastava14a}. 

The network is further adjusted by changing max-pooling to convolutional layers with stride 2. This fully convolution approach should help for later introspection while keeping the same performance \cite{DBLP:Springenberg}. 

While the original network is trained for the 1,000 classes of the ImageNet dataset, here only 4 output units are required for the classes: Berlin, Madrid, Paris and Other. Because of the few classes, also the fully connected layer sizes are reduced from 4096 to 256 which leads to a complexity reduction of 99\% from $220,971,520$ to $66,560$ connections.\footnote{2 layers with size $4096 = 2^{12}$ plus the output layer $1000 \approx 2^{10}$ result in $2^{24} + 2^{22} = 20,971,520$. Whereas 2 layers with size $256 = 2^{8}$ plus the output layer $4 = 2^2$ result in $2^{16} + 2^{10} = 66,560$.} This results into a significant faster training round-trip.

\subsubsection{Training}

For this experiment the Other class consists of all images available. This approach is taken, because it is usually the case that collected data is unlabelled in the first place. The idea is still to find with only a single labelled image for a city additional instances within the relative small, noisy and unlabelled dataset. 

Therefore the Other class consists of 3,126 images including the single reference augmentation samples for the specific cities and further images of these classes with a total of 56 for Berlin, Madrid and Paris. Taken each image 20 variants are produced for Other as described in Section \ref{section:dataset} resulting into 62,520 images containing $56 \cdot 20 = 1120$ ones that are actually part of the classes Berlin, Madrid and Paris. Nevertheless, these represent only about 0,02\% of the Other class. Therefore the assumption is that these samples are acting as natural noise within the dataset.

The single reference augmentation images are shuffled and split into a training and validation set of 50,000 and 12,520 images. Accordingly for Berlin, Madrid and Paris the same amount of images is produced and split using only a single city image respectively. The split into an additional validation set should show evidence during training that not the specific images within the set are learned.

The training procedure has been run with three different configurations each over 50 epochs with a batch size of 64 images:

\begin{enumerate}[label=(\Alph*)]
	\item Optimization of an unregularized cross entropy loss with Adam and in-active dropout layers
	\item Optimization of a L2-regularized cross entropy loss with Adam and in-active dropout layers
	\item Optimization of a L2-regularized cross entropy loss with Adam and active dropout layers
\end{enumerate}

The expected outcome for active dropout layers in (C) is less overfitting on the training dataset and therefore a higher recall during evaluation. A regularization approach as in (B) is in particular necessary as the training dataset is created from only a single reference image per class. Without any regularization as in (A) the assumption is that network will easily overfit on the training data.

The configurations without dropout show already after the first epoch 98\% accuracy on the validation set (Appendix \ref{app:cnn_acc}). The dropout configuration achieves this accuracy only after two epochs. Then the accuracy values are stabilized around 98\%. Only the unregularized configuration shows a massive drop in accuracy between epoch 22 and 23. After epoch 25, the unregularized configuration breaks down and the accuracy stays on values around 25\% which is random choice for 4 classes.

Despite of the high accuracy already after the first epoch, the regularized configurations keep improving the loss until epoch 25 (Appendix \ref{app:cnn_loss}). After that the loss is converged on a value between 0.02 and 0.05 for both. The unregularized configuration is already converged to values under 0.05 after half of these epochs and breaks around epoch 25 with a constant loss of 1.4 afterwards. For all configurations the loss curves show spikes which are in general lowest for the dropout configuration.

\subsubsection{Evaluation}

For the evaluation an unbalanced labelled test dataset is created from all images. The city classes are represented with 12 Berlin, 13 Paris and 31 Madrid images. The Other class is represented with 3,063 images. There are two Berlin, three Paris and two Madrid images removed from the overall dataset which were mislabelled. The test dataset does not contain transformed images, but preprocessed originals as described in Section \ref{section:dataset}.

The trained CNN is applied on the test dataset to predict the four classes Berlin, Paris, Madrid and Other. The prediction results consists of the probability per-class for each image. The actual predicted class is  the class with the highest probability. Given the prediction results the metrics \textit{accuracy, precision, recall and the confusion matrix} are computed. The confusion matrix is further analysed for gathering the \textit{per-class precision and recall}. Finally, for each class the top predicted images are plotted.

The evaluation procedure has been processed after each of the 50 epochs, but is here shown only for the first 25 epochs, because the unregularized configuration is only comparable to the other configurations before the breakdown.

\subsubsection{Results}

The evaluation results are computed including the Other class that represents 98\% of the test images. Therefore the mean precision and recall over all classes are robust to changes in Other, but sensitive to changes in Berlin, Paris and Madrid which are in total only 56 images. This is also the reason why precision and recall are evaluated directly and why accuracy is ignored. The accuracy is always around the same high level without revealing useful properties.

\begin{figure*}[t]	
		\centering
		\subfloat
		{
			\includegraphics[draft=false,width=0.45\textwidth]{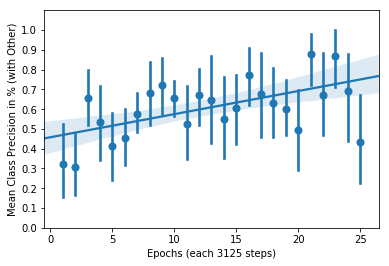}
		}
		\hfill
		\subfloat
		{
			\includegraphics[draft=false,width=0.45\textwidth]{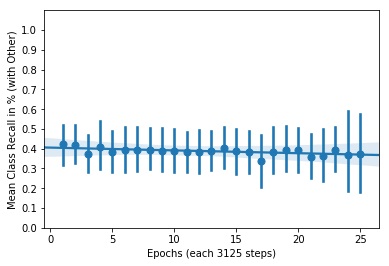}
		}
		\caption{Overall mean precision and recall for all modes and classes per epoch. The precision is the mean of the precisions for all four classes including Other. Accordingly the mean recall is computed.}
		\label{fig:overall_results}
	\end{figure*}

The overall evaluation results as shown in Figure \ref{fig:overall_results} show an increasing precision starting from 30\% at the first epoch to up to 80\% in epoch 23. Then there is a drop in overall precision at epoch 25 towards 40\%. The overall recall is consistently around 40\%.

\begin{figure*}[b]	
		\centering
		\subfloat
		{
			\includegraphics[draft=false,width=\textwidth]{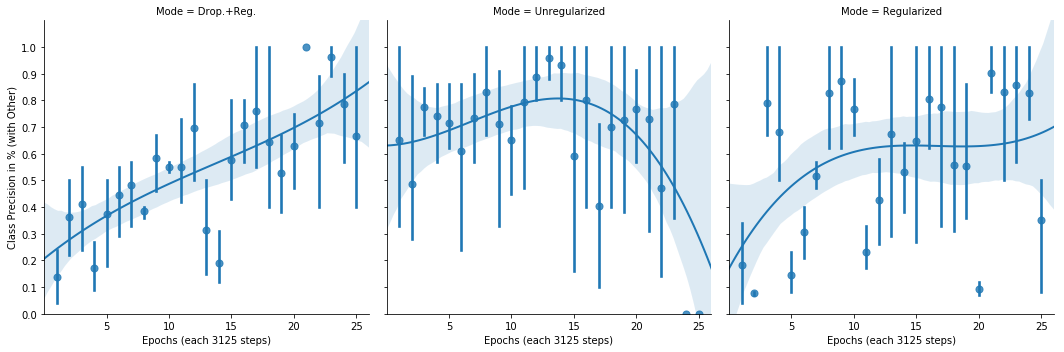}
		}
		\hfill
		\subfloat
		{
			\includegraphics[draft=false,width=\textwidth]{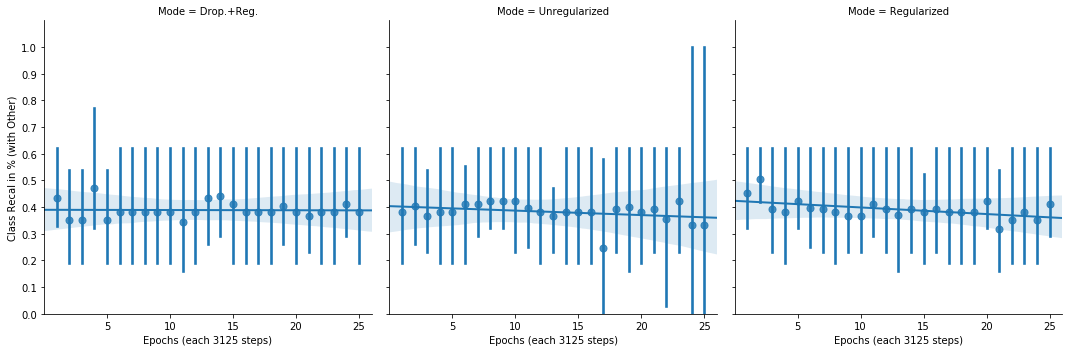}
		}
		\caption{The mean precision and recall for all classes per epoch separated by configuration. The precision is the mean of the precisions for all four classes including Other. Accordingly the mean recall is computed. A third order polynomial is fitted through the precision data points.}
		\label{fig:permode_results}
	\end{figure*}

The per configuration recall as depicted in Figure \ref{fig:permode_results} shows the same consistency for across all configurations with values between 30\% and 50\%. In contrast to that the precision curves show differences among the configurations:
\begin{enumerate}[label=(\Alph*)]
	\item The unregularized configuration starts already on a high precision around 65\%, then the precision is improving until epoch 15 epoch towards 90\%. After that the precision drops with finally a breakdown on epoch 24. 
	\item The regularized configuration starts with low precision around 20\%, then increases towards a mean value around 70\% on epoch 15 and then plateauing with some uncertainty because of frequent outliers. 
	\item The most stable increase in precision is visible for the regularized configuration with dropout. The precision starts low around 20\% and then steadily increases towards a mean precision of 80\% in epoch 24 while having only a few outliers.
\end{enumerate}
For all configurations the precision drops in epochs in which the recall is higher.

The per class results for each configuration reveal that the regularized configuration has the highest variance in all class precisions among all configurations, whereas the regularized configuration with dropout shows a clear tendency of increasing precision on continuous epochs. The unregularized configuration results are a mixture of both (Appendix \ref{app:cnn_prec}). 

\begin{enumerate}[label=(\Alph*)]
	\item Regarding precision, the unregularized configuration results show a declining tendency for Berlin from 70\% to 30\%, a small increasing tendency for Paris from 60\% to 100 \% and a rather consistent for Madrid between 80\% and 100\%.
	\item The regularized configuration shows for Berlin and Paris a rather uncertain precision with high variance and an increasing tendency for Madrid from 30\% to 100 \% with further epochs. 
	\item The results for the configuration with dropout are again showing the most consistent tendency for increase in precision along all classes starting around 20 \% and heading to 90\% on continuous epochs.
\end{enumerate}

\afterpage{\clearpage}
	
\afterpage{\clearpage}

\newpage

All configuration results show a similar recall for each class: Berlin recall is between 30\% and 40\%, Paris recall is around 60\% and Madrid recall is most varying between 20\% and 40\%. The recall of 100 \% for Berlin and 0\% for Paris and Madrid in epoch 24 of the unregularized configuration reveal the breakdown (Appendix \ref{app:cnn_rec}).

\subsubsection{Discussion}

All configurations show a relative high precision after 20 epochs between 60 and 70\%. Nevertheless, the most robust learning is presented by the configuration with regularization and dropout. Therefore this configuration is discussed here in more detail.

The dropout configuration has highest precision for the classes Berlin, Parid and Madrid in epoch 21. The confusion matrix after this epoch is shown in Table \ref{table:confusion}.

\begin{table}[h]
	\centering
	\small
	\begin{tabular}{| l | r | r | r | r | }
		\hline
		& Pred. B & Pred. M & Pred. O & Pred. P \\
		\hline
		Berlin & 3 & 0 & 9 & 0 \\
		Madrid & 0 & 7 & 24 & 0 \\
		Other & 0 & 0 & 3063 & 0 \\
		Paris & 0 & 0 & 5 & 8 \\
		\hline
	\end{tabular}
	\caption{The confusion matrix for the regularized dropout configuration after epoch 21. The precision is 100\% for the classes Berlin, Madrid and Paris.}
	\label{table:confusion}
\end{table}

\begin{figure*}[p]	
	\centering
	\subfloat[The single reference augmentation template for Berlin. ]
	{
		\includegraphics[width=0.35\textwidth]{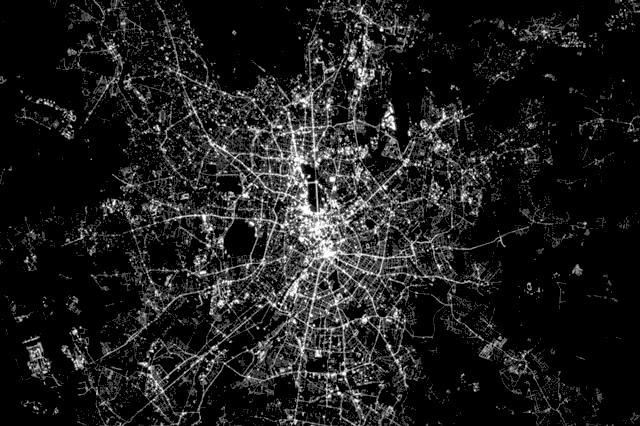}
	}
	\hfill
	\subfloat[The predictions for Berlin in epoch 21 of the dropout configuration.]
	{
		\includegraphics[width=0.6\textwidth]{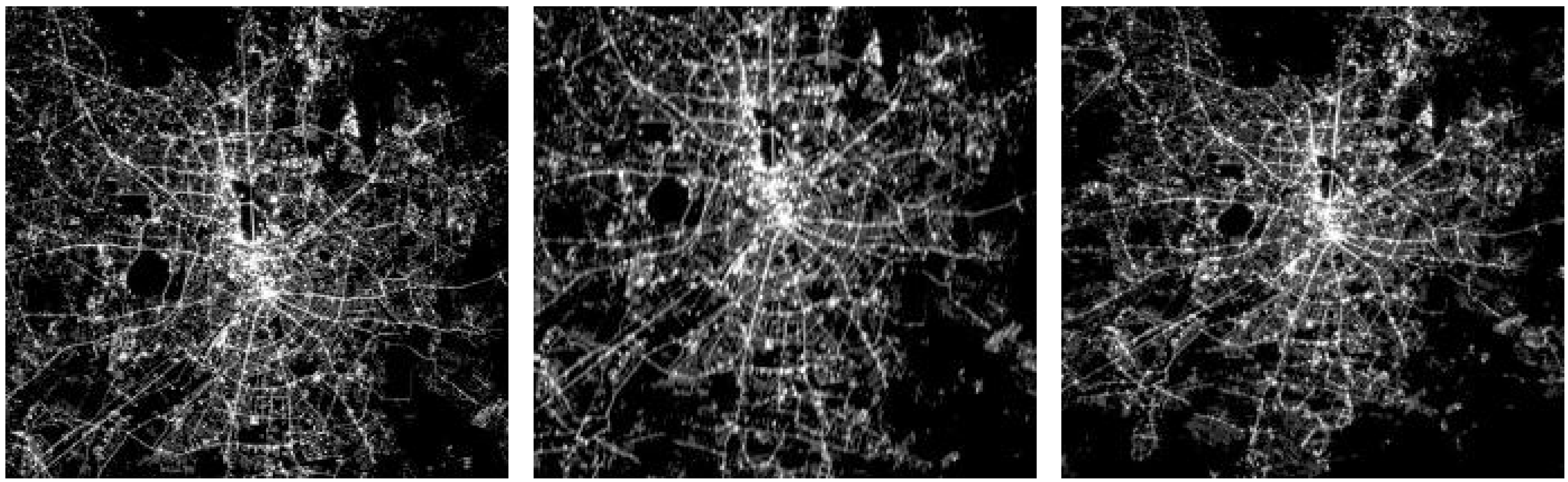}
	}
	\hfill
	\subfloat[The single reference augmentation template for Madrid.]
	{
		\includegraphics[width=0.35\textwidth]{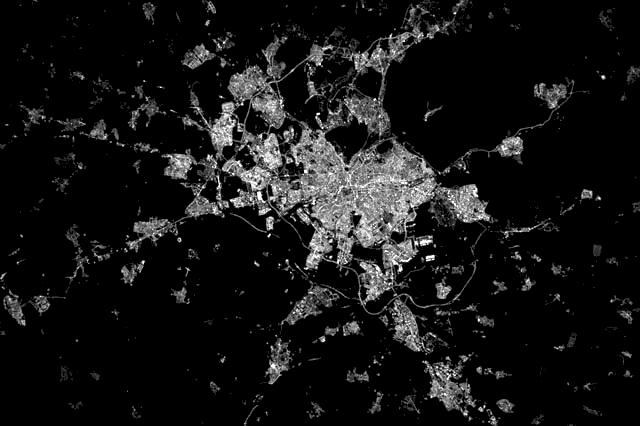}
	}
	\hfill
	\subfloat[The predictions for Madrid in epoch 21 of the dropout configuration.]
	{
		\includegraphics[width=0.6\textwidth]{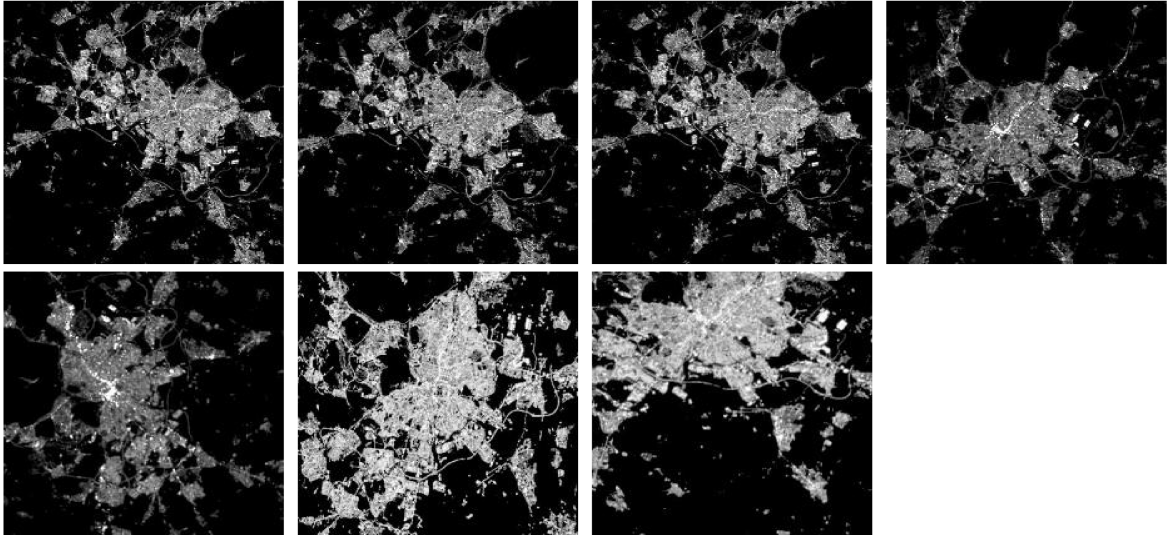}
	}
	\hfill
	\subfloat[The single reference augmentation template for Paris.]
	{
		\includegraphics[width=0.35\textwidth]{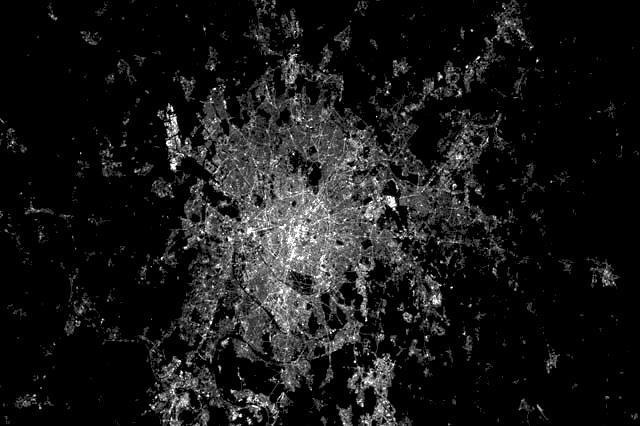}
	}
	\hfill
	\subfloat[The predictions for Paris in epoch 21 of the dropout configuration.]
	{
		\includegraphics[width=0.6\textwidth]{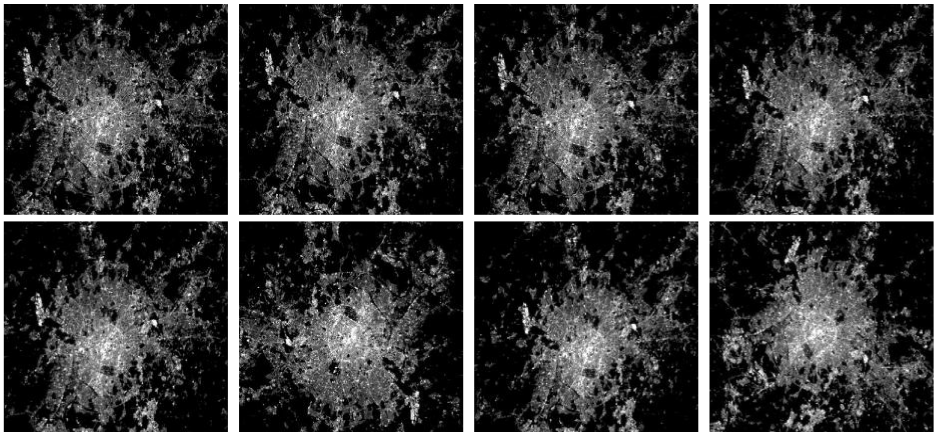}
	}
	\caption{The single reference augmentation references for each class and the predictions for the dropout configuration in epoch 21. The template images are enhanced originals of size 640x426, because only the created transformed image based on the template is cropped. }
	\label{fig:predictions_epoch21}
\end{figure*}

The according predictions in Figure \ref{fig:predictions_epoch21} show that the dropout configuration is capable of predicting images with high precision within the noisy, unclean and unbalanced \textit{Cities at Night} dataset that a human classifier would describe as similar to the corresponding single reference augmentation templates. The predicted images are variants of their templates in shift, zoom and rotation. The predictions for Madrid show also a rather detailed image of the city. 

\begin{figure*}[p]
	\centering
	\includegraphics[width=\textwidth]{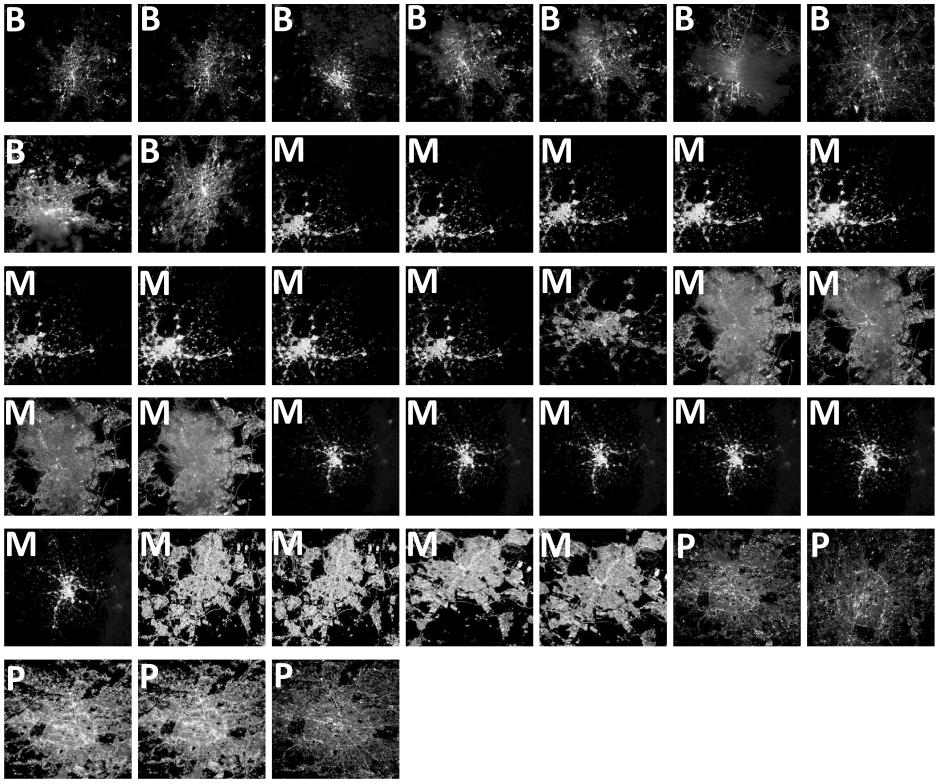}
	\caption{The predictions for Other in epoch 21 that contain images of the labelled cities. The letter in the upper left corner signals the correct class: B = Berlin, M = Madrid, P = Paris.}	
	\label{fig:predictions_other_epoch21}
\end{figure*}

In contrast to this, the predictions of labelled city classes as Other as shown in Figure  \ref{fig:predictions_other_epoch21} are rather different from the chosen template image. The dropout configuration after epoch 21 has learned a useful embedding to identity variants of the templates, but has not learned useful patterns to identity rather different examples of the cities:

\begin{enumerate}[label=(\alph*)]
	\item The highest recalls for Berlin are after epoch 13, 14 and 15 with 42\% and 5 out of 12 correctly predicted images, but also 5 images from Other. Therefore the precision is only 50\%.	
	\item The highest recall for Madrid is after epoch 1 with 35\% and 11 out of 31 correctly predicted images, but also 34 images from Other. Therefore the precision is only 24\%.
	\item 	The highest recall for Paris is after epoch 4 with 10 out 13 predicted images, but also 104 of other classes as Paris. Therefore, although the recall is here 77\%, the precision here is only about 9\%.
\end{enumerate}

\begin{figure*}[p]
	\centering
	\includegraphics[width=\textwidth]{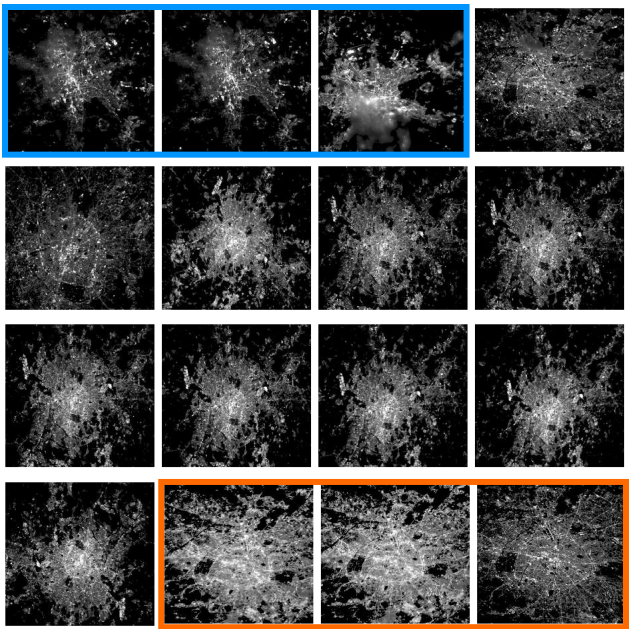}
	\caption{The matching predictions for Paris among the classes for the dropout configuration which results show a high recall, but low precision, after epoch 4. The orange boxed images are actually Paris images but predicted as Other. The blue boxed images are predicted as Paris, but show actually Berlin. The Paris predictions of Other images are not shown.}	
	\label{fig:predictions_recall_paris}
\end{figure*}

The predictions shown in Figure \ref{fig:predictions_recall_paris} underline the assumption that the trained convolutional neural network is unfortunately not capable to achieve a high recall while keeping precision, when the Other class is overrepresented. This effect might be strengthened by the fact that actual labelled city classes are within the Other class during training.

Nevertheless, when the network is run for a high recall on the noisy, unclean and unbalanced dataset, the dropout configuration shows capabilites to highly reduce the amount of cities that would need to be checked by a human labeller while still finding new variants of the city.

Therefore the network could be run in an automatic tooling environment with high precision and for further introspection of a large \textit{Cities at Night} dataset for high recall. In the end, the effect of loosing precision on high recall might be tackled by adding more classes so that the CNN has to learn more distinguishing features from the images.

\section{Conclusion and Further Work}

This paper demonstrates that \textit{Single Reference Augmentation} creates useful training datasets for deep learning techniques, when the transformation results are coherent with the image variants found in the original dataset. Then the stated technique is useful for application, when there are only a few samples within the dataset per category.

The findings are that a convolutional neural network, that has been trained on such a dataset, is capable of finding images of the same category in a noisy, unclean and unbalanced dataset, whereas the applied principal component analysis in conjunction with a Fourier transform failed to solve this task. 

The results have shown that the convolution neural network easily finds images that are similar the single reference, but that it struggled to find more different images while keeping the same high precision.

Further work would include introspection of the convolution neural network to analyse which patterns were found and if these are useful or artificial. Based on this, methods to increase the recall while keeping precision need to be researched. This problem could also be approached by adding more categories to the training, so that the network is driven to find more distinguishing features from the images.

\bibliography{naaclhlt2016}
\bibliographystyle{naaclhlt2016}

\onecolumn
\begin{appendices}
	
	\newpage
	\section{Cities at Night dataset examples}
	\label{appendix:quality}
	
	\begin{figure}[H]
		\centering
		\subfloat[Several images in the dataset are showing the earth ellipse.]
		{
			\includegraphics[width=0.5\textwidth]{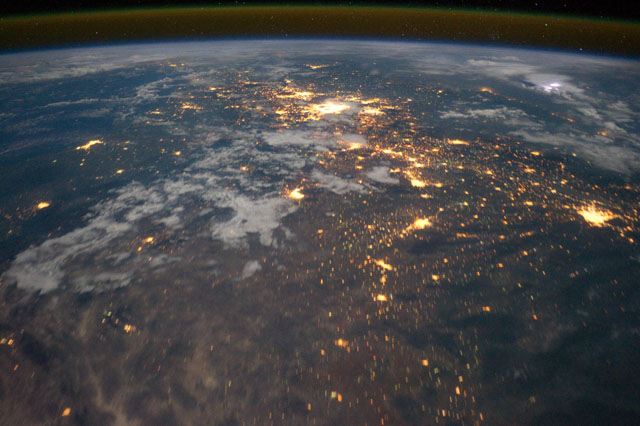}
		}
		\hfill
		\subfloat
		[Many images in the dataset show multiple cities and are in parts blurry.]
		{
			\includegraphics[width=0.5\textwidth]{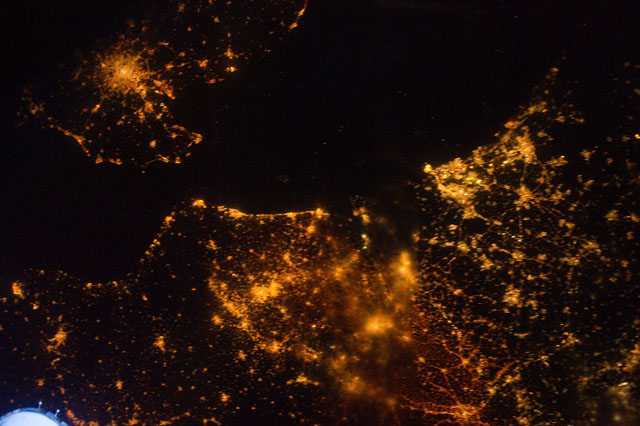}
		}
		\hfill
		\subfloat
		[Some images in the dataset contain parts of the station.]
		{
			\includegraphics[width=0.5\textwidth]{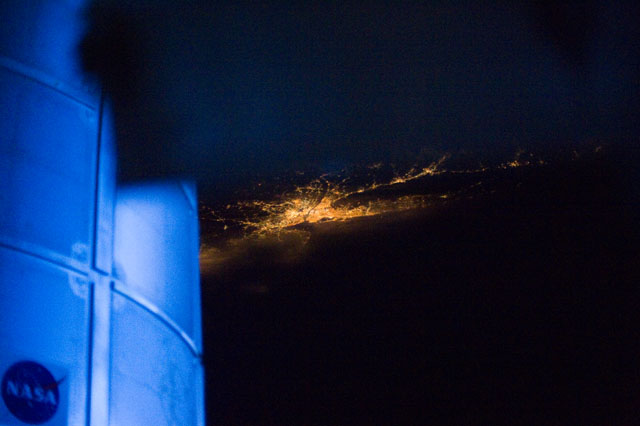}
		}
		\caption{The quality of the images within the Cities at Night dataset is highly varying as shown here example-wise.}
		\label{fig:quality}
	\end{figure}
	
	\section{CNN training accuracy until epoch 20}
	\label{app:cnn_acc}

	\begin{figure}[H]	
		\centering
		\subfloat[Accuary for an unregularized loss without dropout.]
		{
			\includegraphics[width=\textwidth]{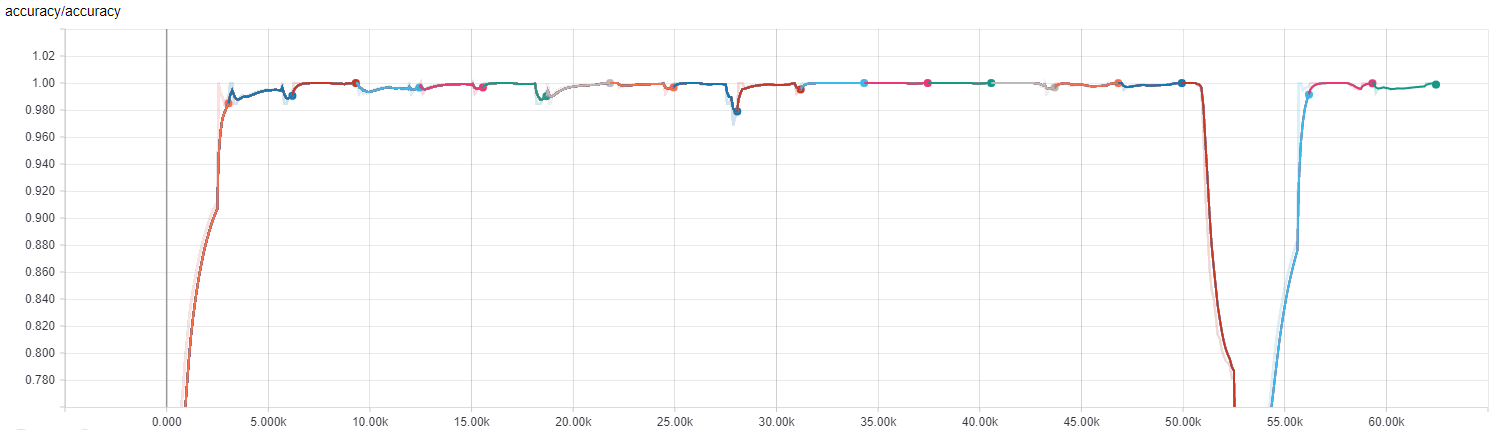}
		}
		\hfill
		\subfloat[Accuary for a regularized loss without dropout.]
		{
			\includegraphics[width=\textwidth]{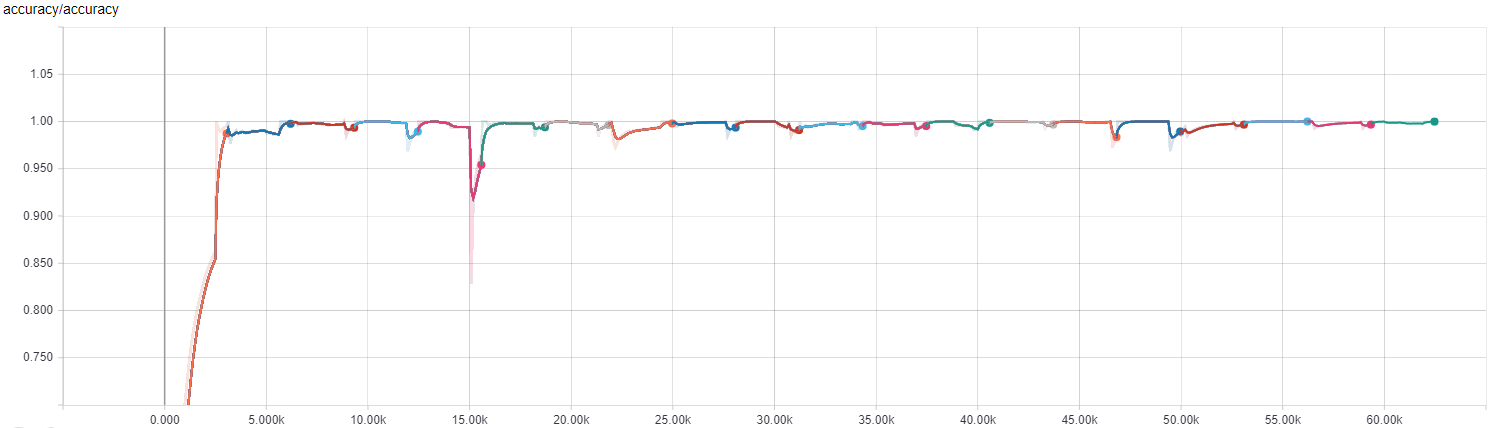}
		}
		\hfill
		\subfloat[Accuary for a regularized loss with dropout.]
		{
			\includegraphics[width=\textwidth]{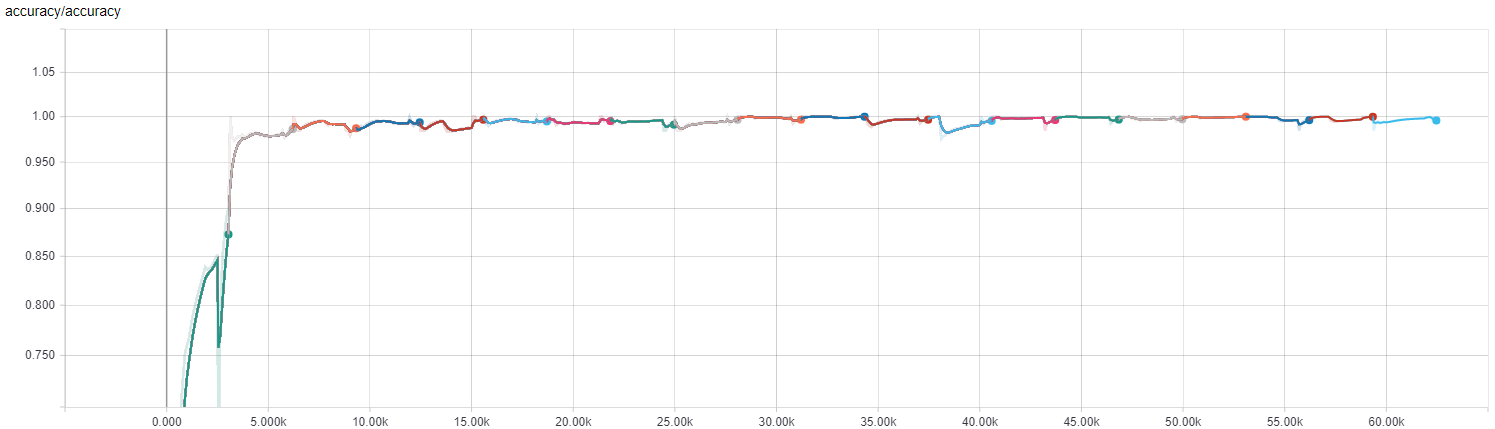}
		}
		\caption{The accuracy on the validation set for all three configurations in the first 20 epochs. The  accuracy is given in percent (1.00 = 100\%) on the y-axis and the number of steps on the x-axis. An epoch is about 3125 steps and indicated by an individual colour. The values are smoothed with 60\%.}
		\label{fig:acc20}
	\end{figure}
	
	\section{CNN training loss until epoch 20}
	\label{app:cnn_loss}
	
	\begin{figure}[H]	
		\centering
		\subfloat[Unregularized loss without dropout.]
		{
			\includegraphics[width=\textwidth]{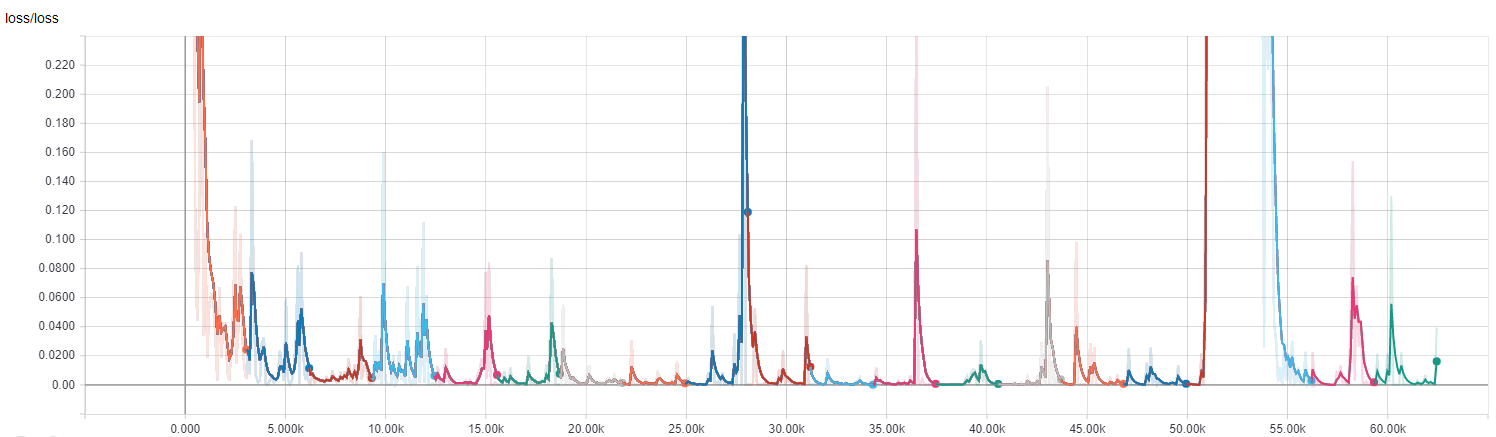}
		}
		\hfill
		\subfloat[Regularized loss without dropout.]
		{
			\includegraphics[width=\textwidth]{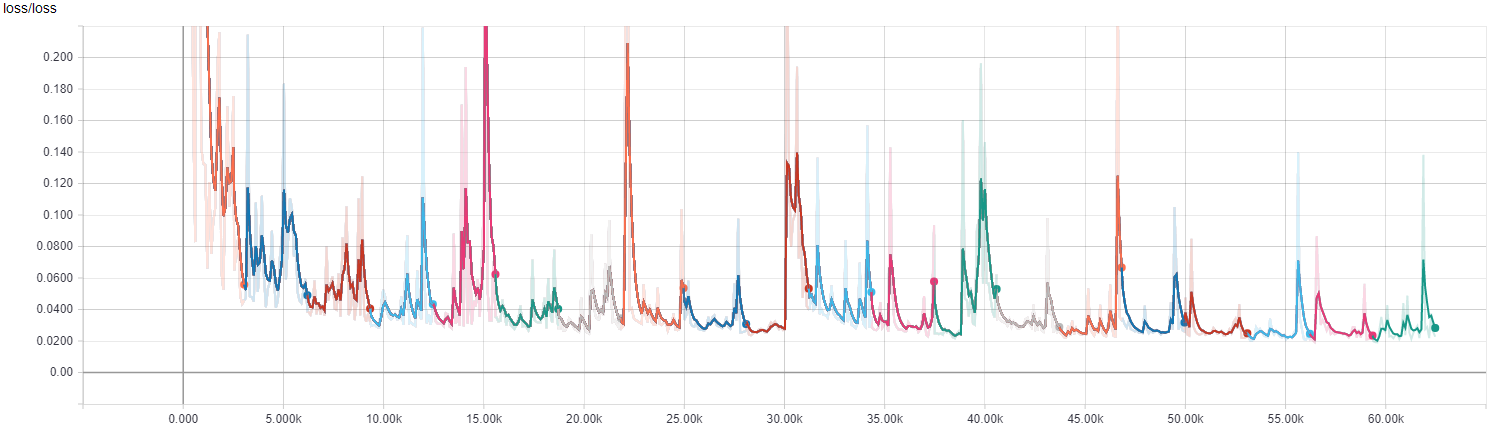}
		}
		\hfill
		\subfloat[Regularized loss with dropout.]
		{
			\includegraphics[width=\textwidth]{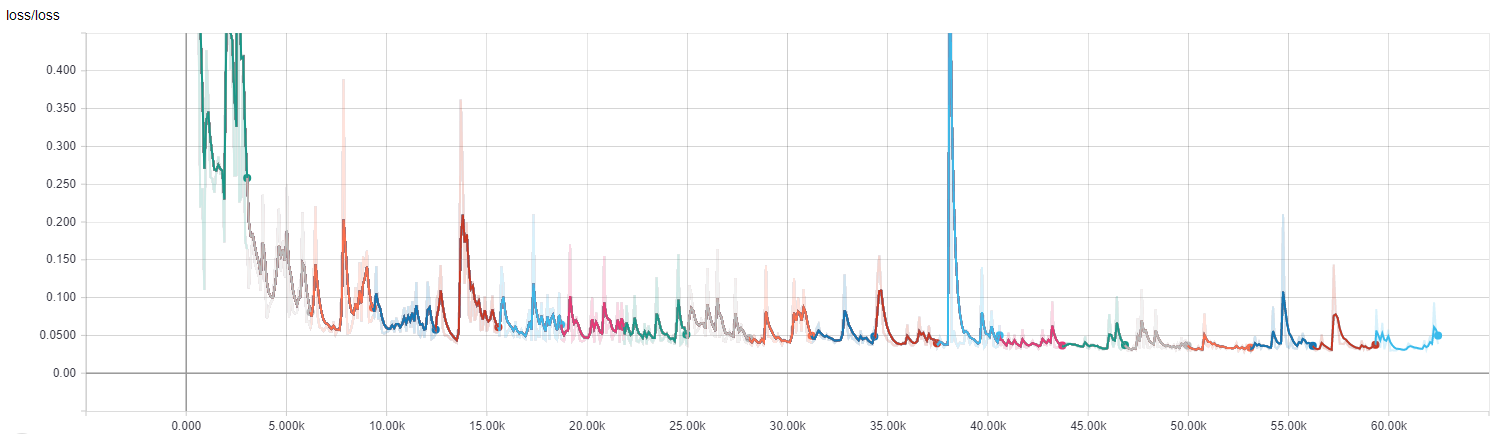}
		}
		\caption{The loss curves for all three configurations in the first 20 epochs. The total loss is given on the y-axis and the number of steps on the x-axis. An epoch is about 3125 steps and indicated by an individual colour. The values are smoothed with 60\%.}	
		\label{fig:loss20}
	\end{figure}
	
	\section{CNN precision in percent for the first 25 epochs separated by configuration and class}
	\label{app:cnn_prec}
	
	\begin{figure}[H]
		\centering
		\includegraphics[width=\textwidth]{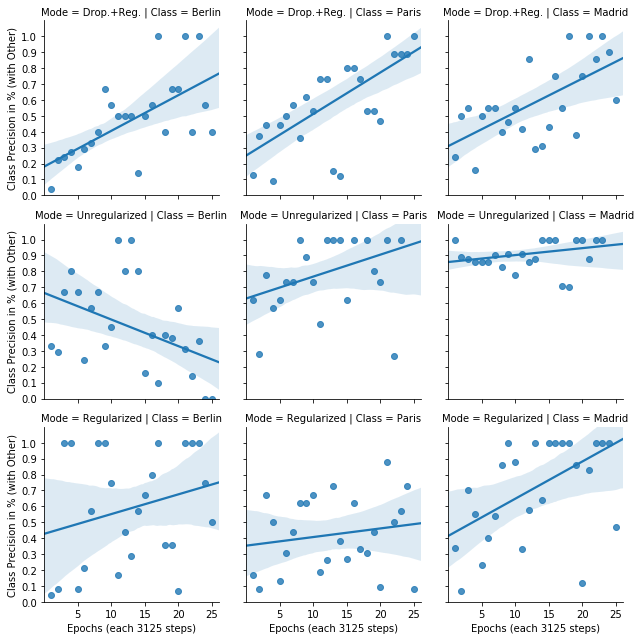}
		\caption{The precision in percent per epoch separated by configuration and class (not showing Other). A linear model is fitted through the data points.}
		\label{fig:perclass_precision}
	\end{figure}
	
	\section{CNN recall in percent for the first 25 epochs separated by configuration and class}
	\label{app:cnn_rec}
	
	\begin{figure}[H]
		\centering
		\includegraphics[width=\textwidth]{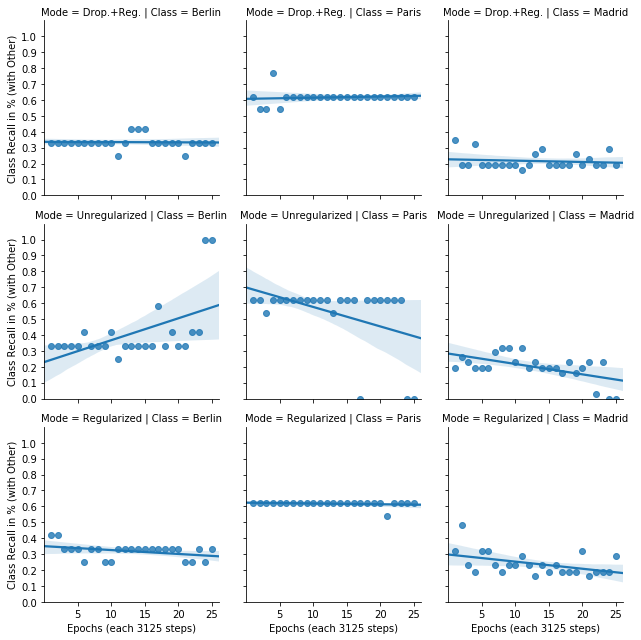}
		\caption{The recall in percent per epoch separated by configuration and class (not showing Other). A linear model is fitted through the data points.}	
		\label{fig:perclass_recall}
	\end{figure}
	
\end{appendices}

\end{document}